\documentclass[11pt]{article}
\usepackage{eamt24}
\usepackage{times}
\usepackage{url}
\usepackage{latexsym}
\usepackage[small,bf]{caption} 

\usepackage[symbol]{footmisc}
\usepackage{amssymb}
\usepackage{booktabs}
\usepackage{amsmath}
\DeclareMathOperator*{\argmin}{argmin^k}
\DeclareMathOperator*{\avg}{avg}
\usepackage{graphicx}
\usepackage{subcaption} 

\title{Quality Estimation with $k$-nearest Neighbors and Automatic Evaluation for Model-specific Quality Estimation}

\author{Tu Anh Dinh$^1$\\
  \normalsize{\tt tu.dinh@kit.edu}
  \And
  Tobias Palzer$^2$ \\
  \normalsize{\tt tobiaspalzer.tp@gmail.com} \vspace{0.5cm} \\
  $^{1,3}$Karlsruhe Institute of Technology \\Karlsruhe, Germany \\
  $^2$Technical University of Munich \\Munich, Germany 
  \And
  Jan Niehues$^3$ \\
  \normalsize{\tt jan.niehues@kit.edu} 
  }

\date{}

\begin{document}
\maketitle
\footnotetext{$^{1,2}$Equal contributions.}
\footnotetext{$^2$Tobias contributes during his Bachelor Thesis at KIT.}
\begin{abstract}
Providing quality scores along with Machine Translation (MT) output, so-called reference-free Quality Estimation (QE), is crucial to inform users about the reliability of the translation. We propose a model-specific, unsupervised QE approach, termed $k$NN-QE, that extracts information from the MT model's training data using $k$-nearest neighbors. Measuring the performance of model-specific QE is not straightforward, since they provide quality scores on their own MT output, thus cannot be evaluated using benchmark QE test sets containing human quality scores on premade MT output. Therefore, we propose an automatic evaluation method that uses quality scores from reference-based metrics as gold standard instead of human-generated ones. We are the first to conduct detailed analyses and conclude that this automatic method is sufficient, and the reference-based MetricX-23 is best for the task. 
\end{abstract}

\section{Introduction}






Machine Translation (MT), due to its currently advanced stage in research, has been widely adopted in real-life use cases \cite{vieira2021understanding}. In many application domains such as health care or lawsuits, errors in translation could be tremendously harmful to the users. Therefore, it is important to inform the user whether to rely on a certain translation, by providing some kind of quality assessment along with each translation output. This task is referred to as Quality Estimation (QE). 

More specifically, Quality Estimation is assigning quality scores to MT output, without using gold-standard human translation. 
Common QE approaches train a standalone QE module that takes in the source sentences and the MT outputs to produce quality scores. These QE modules are usually model-agnostic, i.e., they can work with the output of any MT model. However, they often require training on human-labeled quality data, which can be costly to obtain.
Another line of research is on model-specific QE, where they exploit or modify the MT model for self-quality assessment, thus not requiring training a separate QE module.
Following this line of work, we propose \textit{$k$NN-QE} - an unsupervised QE approach that exploits the information of inference-time output's $k$-nearest neighbors found in the MT model's training data. We hypothesize that the closer the inference-time sample output is to the training data, the better the quality of the translation, since it is an indication that the model has learnt about such samples. The QE scores obtained using our method can also be interpreted as the confidence scores of the MT model.

Unlike model-agnostic QE approaches which can take any MT translation as input, evaluating model-specific QE approaches like $k$NN-QE is not as straightforward. Public QE test sets are generated using human quality scores on pre-made MT output, thus not always suitable for QE approaches that perform self-evaluation on their own MT output by design. Many previous works on model-specific QE perform human evaluation on their own MT output to be used as gold standard to evaluate QE metrics \cite{rikters-fishel-2017-confidence,niehues-pham-2019-modeling,fomicheva-etal-2020-unsupervised,zhang-etal-2022-competency}. However, for faster development, it would be useful to automatically evaluate QE metrics and not relying on human resource. Therefore, we propose using quality scores generated by reference-based metrics as the gold standard to automatically evaluate reference-free QE metrics. Our motivation is that reference-based metrics, by making use of reference translation, tend to be better than reference-free QE metrics \cite{freitag-etal-2023-results}, thus can be used as gold standard. To the best of our knowledge, we are the first to perform a detailed analysis on whether reference-based metrics are sufficient to evaluate QE, and which reference-based metric is best suited. For this analysis, we make use of different QE submissions to public shared tasks: WMT22 Metrics \cite{freitag-etal-2022-results} and WMT22 Quality Estimation \cite{zerva-etal-2022-findings}. We investigate whether our automatic QE evaluation method can produce similar QE rankings compared to using human-labeled quality data in these shared tasks.

In summary, our contribution is in two folds:
\begin{enumerate}
    \item A \textbf{model-specific, unsupervised QE} approach, termed \textit{$k$NN-QE}\footnote{https://github.com/TuAnh23/auto-meta-eval-qe}, which exploits the similarity of MT generated output and MT models' training data. Our main findings are: (1) $k$NN-QE outperforms an unsupervised baseline using MT output probabilities, but falls behind supervised QE; and (2) $k$NN-QE works with a small number of neighbors and partial access to the MT training data.
    
    \item An \textbf{automatic QE evaluation} method\footnote{https://github.com/TuAnh23/knn-box} using a reference-based metric's quality scores as gold-standard instead of human-labeled quality scores. Our main findings are: (1) QE ranking made by reference-based metrics correlate well with ones made by human quality scores; (2) Segment-level evaluation performance does not strictly correlate to QE ranking performance for reference-based metrics; and (3) MetricX 23 \cite{juraska-etal-2023-metricx} is the most robust for ranking QE metrics.
\end{enumerate}

\section{Related Work} \label{sec:related_work}
\paragraph{Quality Estimation} Quality Estimation (QE) aims to measure the quality of MT output without using human references. 
Common QE approaches are model-agnostic, where a QE module takes in a source sentence, an MT translation and outputs a quality score \cite{blain-etal-2023-findings}. This approach has 2 drawbacks: (1) it requires a stand-alone module for QE, and (2) it requires human quality data to train the QE module, which can be costly.

\paragraph{Model-specific QE} Researchers have also been looking into integrating Quality Estimation into MT models. These approaches exploit information or modify white-box MT model to measure the translation quality, rather than training a separate QE module relying completely on human quality data. \cite{rikters-fishel-2017-confidence} uses the attention distributions from the MT model as a QE metric. \cite{fomicheva-etal-2020-unsupervised} uses the attention distribution and the output probabilities from the MT model for QE. \cite{lu-etal-2022-learning} propose QE learnt jointly with the training of the MT model. In their approach, the MT model can ask for hints to improve its translation, and the more hints it asks for, the lower the confidence. \cite{zhang-etal-2022-competency} extends the MT model with a self-estimator module for QE, which examines whether it can reconstruct the source sentence' semantics using the information from the decoding procedure. The work by \cite{niehues-pham-2019-modeling} is the closest to our \textit{$k$NN-QE}, where they measure the similarity of the test sentence with sentences from the training data to estimate translation quality. The difference between this work and ours is that they focus on evaluating source side rather than target side; they use encoder output similarity rather than decoder output similarity; and they do not analyze different metrics derived from the nearest neighbors. Evaluating these model-specific QE approaches is not straightforward, as will be discussed below.

\paragraph{Automatic QE evaluation} The standard way to evaluate QE is to use some benchmark test sets, containing human quality scores on the output of some MT models. The QE scores are then compared against the human scores on these pre-made translations. This works mostly for model-agnostic QE, since they can evaluate any MT output. However, for the model-specific QE approaches like $k$NN-QE, which provide quality scores on their own MT output, there are no longer readily available human quality scores for QE evaluation. 
Previous works on model-specific QE address this issue differently. Some works use MT glass-box features for QE without changing the MT model, thus they can still produce the same MT translation that is used in the QE benchmarks \cite{yankovskaya-etal-2018-quality,wang-etal-2021-beyond-glass}. \cite{lu-etal-2022-learning} train their MT model on the same data as the model used in the QE benchmarks and perform force decoding to get the exact same MT output. These approaches are then limited to the MT model used in the QE benchmarks. On the other hand, some works perform human evaluation on their own MT output for QE evaluation \cite{rikters-fishel-2017-confidence,niehues-pham-2019-modeling,fomicheva-etal-2020-unsupervised,zhang-etal-2022-competency}. This requires human resource, which is costly and not always available. Overall, it is not yet clear what is the go-to method to perform automatic evaluation for model-specific QE. To the best of our knowledge, we are the first to perform detailed analysis on whether it is possible to automate evaluation for QE by making use of reference-based metrics.

\paragraph{$k$NN for generation tasks} Previous works have applied $k$-nearest neighbors in text generation. $k$NN-LMs \cite{khandelwal2019generalization} enable language models to interpolate their token prediction output with a $k$-nearest neighbors model, where nearest neighbors are retrieved from a datastore of sample representations.
$k$NN-MT \cite{khandelwal2020nearest} also enables the MT model to predict tokens using a nearest neighbor classifier over a datastore of representations. $k$NN-LMs and $k$NN-MT are particularly useful for adapting models to diverse domains by using domain-specific datastores. 
Our $k$NN-QE approach is similar to these works in two aspects. First, in the datastore generation process, it also generates token representations by performing one forward pass of the model through the training data. Second, during inference, it also retrieves similar tokens in the datastore based on the token representation distance. The difference is that our $k$NN-QE approach uses the retrieved neighbors to assess the quality of the generated token, rather than modifying the model output like $k$NN-LMs and $k$NN-MT.

\section{Quality Estimation with $k$NN}

\begin{figure*}[t]
\centering
\includegraphics[width=0.9\textwidth]{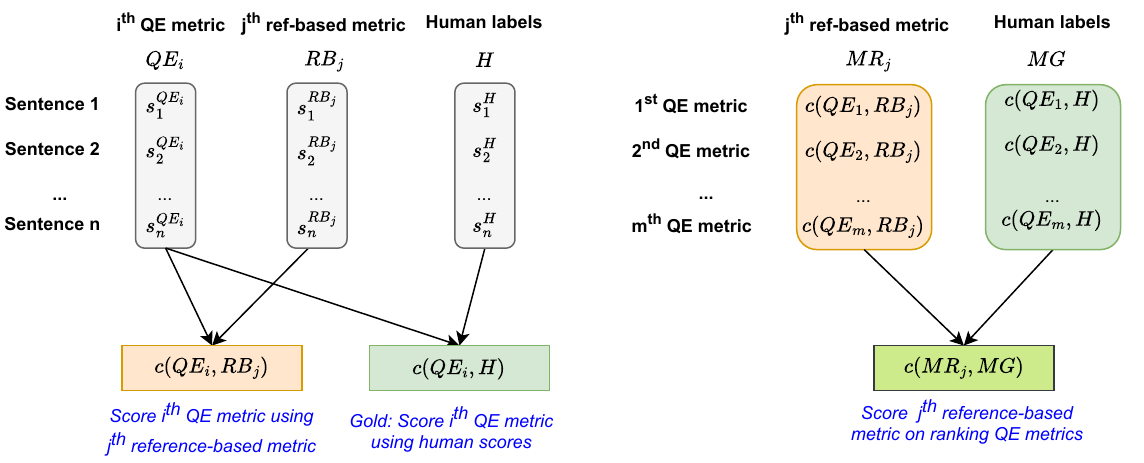}
\caption{Illustration of our automatic QE evaluation approach.}
\label{fig:QEEE}
\end{figure*}

\paragraph{Motivation} We propose \textit{$k$NN-QE} - a model-specific Quality Estimation method that exploits information from the MT model's training data using $k$-nearest neighbors. Our method is unsupervised, thus does not require human quality scores for training. Generally, if the hidden representation of a translation sample generated during inference is similar to ones generated on the training data, then it is an indication that this sample is in-domain, thus more likely to have higher quality.

\paragraph{Generating the datastore} We generate a datastore on the MT training data as follows. We first use the MT model to perform translation on its training set with force decoding on the reference. That is, we give the model human reference translation prefixes as input at every time step to generate the next translation token. We save the last-layer decoder hidden representation of every output token to the datastore. We do forced decoding on the reference for datastore generation since it provides an indication of confidence: if during inference, the self-generated prefix translation is high-quality, it would better match the forced decoding condition where prefixes are gold translation, thus making the representation of the inference-time generated token closer to the ones in the datastore. 

Formally, let the $m^{th}$ training source sentence be $X^m=(x_1^m, x_2^m,..,x_{|X^m|}^m)$ and the $m^{th}$ training reference target sentence be $\hat{Y}^m=(\hat{y}_1^m, \hat{y}_2^m,..,\hat{y}_{|\hat{Y}^m|}^m)$, where the element tokens are subwords. The last-layer decoder hidden representation of the output token at time step $i$ with forced decoding on the reference is:
\begin{equation}
\hat{d}_i^m = Dec(E^m, (\hat{y}_1^m, \hat{y}_2^m,..,\hat{y}_{i-1}^m))
\end{equation}
where $E^m = Enc(X^m)$, $Dec$ and $Enc$ are the decoder and encoder functions respectively. We save to the datastore the $\hat{d}_i^m$ representation for each output token $\hat{y}_i^m$ in the training data.

\paragraph{Retrieving neighbors during inference} During inference, for each generated token, we use its last-layer decoder hidden representation and find the $k$-nearest neighbors from the datastore. The neighbor retrieval can be highly optimized using toolkits like Faiss \cite{johnson2019billion}, thus does not cost too much inference speed.

Formally, let the output sentence be $Y=(y_1, y_2,..,y_{|Y|})$. The last-layer decoder hidden representation of the output token $y_j$ at time step $j$ is (no forced decoding at inference time):
\begin{equation}
d_j = Dec(E, y_1, y_2,..,y_{j-1}))
\end{equation}
where $E = Enc(X)$. We find the set $N_j$ of $k$-nearest neighbors of $y_j$ by:
\begin{equation}
N_j = \argmin_{m\in \text{train}, i \in 1..|\hat{Y}^m|}(L_2(d_j, \hat{d}_i^m))
\end{equation}
where $\argmin$ returns the indices of $k$ smallest elements and $L_2$ is the Euclidean distance.

\paragraph{Derive QE metrics} Given the retrieved $k$-nearest neighbors, we derive QE metric $s_j$ for each inference-time generated token $y_j$:
\begin{itemize}
    \item $k$NN token distance: We calculate the average distance from $y_j$ to its $k$-nearest tokens in the datastore:
    \begin{equation}
    s_j=
    \avg_{(m,i) \in N_j}(L_2(d_j, \hat{d}_i^m))
    \end{equation}
    We assume the lower the distance, the better the translation quality, since the generated token is familiar to the MT model.
    \item $k$NN sentence similarity: We calculate the average cosine similarity between the whole inference-time generated sentence and the K sentences in the training data to which the $k$NN tokens belong:
    \begin{equation}
    s_j = \avg_{(m,i) \in N_j}(
    cos\_sim(emb(Y), emb(\hat{Y}^m))
    )
    \end{equation}
    where $cos\_sim$ is the cosine similarity function, $emb$ is the sentence embedding function. For sentence embedding, we use an external model instead of the MT model itself, since the external model won’t be affected by artifacts in the MT training data.\\
    We assume the higher the similarity, the better the translation quality, since the generated sentence is familiar to the MT model.
    \item Number of different $k$NN tokens: We count the number of distinct tokens amongst the retrieved $k$NN tokens:
    \begin{equation}
    s_j = |\{\hat{y}_i^m | (m,i) \in N_j \}|
    \end{equation}
    We assume the higher the number, the lower the translation quality, since it means the neighbor cluster is not representative of any specific token, indicating that the model is uncertain about the generated representation.
    \item Model prediction equals retrieved $k$NN tokens: We count the number of retrieved $k$NN tokens that are the same as the model output token $y_j$:
    \begin{equation}
    s_j = |[\hat{y}_i^m | (m,i) \in N_j \wedge \hat{y}_i^m = y_j ]|
    \end{equation}
    We assume the higher the number, the better the translation, since it is easy for the model to map the representation to one single token.
\end{itemize}
Using these metrics, we can get quality scores on the token level. To get scores on the segment level, we take the average of the scores of tokens:
\begin{equation}
    s_Y = \avg_{j \in 1..|Y|}(s_j)
\end{equation}

\section{Automatic evaluation for Quality Estimation} \label{sec:meta_eval}

\paragraph{Motivation} Normally, to evaluate QE metrics, people calculate the correlation between QE-generated quality scores and human-generated quality scores on some MT output \cite{zerva-etal-2022-findings}. However, as discussed above, model-specific QE approaches such as $k$NN-QE provide quality scores on their own MT output, thus we cannot evaluate them using the available human quality scores on different MT outputs in the public benchmarks. Collecting human-generated quality scores again on this specific MT output would be costly in terms of time and human resources. Therefore, we propose an automatic approach using reference-based metrics as gold standard to evaluate QE metrics. 

\paragraph{Reference-based metrics as gold} Recall that Quality Estimation takes only the source sentence and the MT translation for outputting a quality score, while reference-based metrics also make use of human gold-standard translation. As a result, reference-based metrics are usually more robust than QE metrics \cite{freitag-etal-2023-results}. Therefore, we attempt to perform automatic evaluation for QE by calculating the correlation between the QE scores and the ref-based metrics scores. In other words, we are using the ref-based metrics scores in place of human-provided scores as the gold standard. We investigate scores at the segment level. Using this approach, we can flexibly generate gold-standard quality scores for any MT output, rather than relying on fixed human quality scores on some pre-made MT output.

\paragraph{Boosting reference-based metrics' reliability} Intuitively, it is important to have a robust reference-based metric since we are using it as gold standard for reference-free QE. One potential way to have more robust reference-based metrics is to increase the number of references. Therefore, we propose to use test datasets with multiple references, and additionally use a paraphraser tool to generate synthetic references. 

\paragraph{Choosing reference-based metric} We investigate whether reference-based metrics are good enough for evaluating QE metrics, and which reference-based metric is best suited. We gather different QE metric submissions on public shared tasks, and measure the correlation between the QE ranking created by human annotations and the QE ranking created by reference-based metrics. An illustration of the process is shown in Figure \ref{fig:QEEE}.

Specifically, assume we have $n$ MT output segments, $m$ QE metrics and $p$ reference-based metrics. Let $\mathrm{QE}_i$, $\mathrm{RB}_j$, $\mathrm{H}$ $\in \mathbb{R}^{n \times 1}$ be the quality scores assigned to the MT translations by the $i^{th}$ QE metric, the $j^{th}$ reference-based metric and the human annotator respectively. The gold evaluation for the QE metrics is then:
\begin{equation}
\mathrm{MG} = (c(\mathrm{QE}_1, H), c(\mathrm{QE}_2, H), ..., c(\mathrm{QE}_m, H))
\label{eq:gold_eval_qe}
\end{equation}
where $c$ is a correlation function such as Spearman. The automatic evaluation for the QE metrics using the $j^{th}$ reference-based metric is:
\begin{equation}
\begin{split}
\mathrm{MR_j} = \hspace{6.3cm} \\
(\mathrm{c}(\mathrm{QE}_1, \mathrm{RB}_j), \mathrm{c}(\mathrm{QE}_2, \mathrm{RB}_j), ..., \mathrm{c}(\mathrm{QE}_m, \mathrm{RB}_j))
\end{split}
\label{eq:ref_eval_qe}
\end{equation}
The performance of the $j^{th}$ reference-based metric on ranking QE metrics is then:
\begin{equation}
c(\mathrm{MR_j}, \mathrm{MG})
\label{eq:rank_qe_eval}
\end{equation}
Note that this is not the same as the performance of the $j^{th}$ reference-based metric on scoring segment-level MT, which is defined as:
\begin{equation}
c(\mathrm{RB}_j, \mathrm{H})
\label{eq:seg_level_eval}
\end{equation}

\section{Experimental Setup}
\subsection{Automatic evaluation for Quality Estimation}
\paragraph{Dataset} In our experiments, we uses the English -- German data from two shared tasks: \textit{WMT22 Quality Estimation} \cite{zerva-etal-2022-findings} and \textit{WMT22 QE as a Metrics} \cite{freitag-etal-2022-results}. The \textit{WMT22 Quality Estimation} shared task, which we refer to as \textit{QE Task}, is specialized in evaluating Quality Estimation. The \textit{WMT22 QE as a Metrics} shared task, which we refer to as \textit{QE-M Task}, is meant for comparing QE metrics to reference-based metrics. Both shared tasks contain submissions from different QE systems. which is useful for us to investigate whether we can automatically rank these QE systems. Specifically, the QE-M Task data includes source sentences, reference sentences and translation sentences from multiple different MT systems from the WMT22 General MT task \cite{kocmi-etal-2022-findings}, along with human-labeled MQM quality score \cite{lommel-etal-2014-using} and QE submission scores on each translation sentence. The data from the QE Task is similar, except that (1) they only use data from the News domain rather than the full test set from the WMT22 General MT task (including the Conversation, Ecommerce, News and Social domains) and (2) the MT output is from a single MT system. More details can be found in Table \ref{tab:meta_eval_qe_data}. 

\begin{table}[htbp]
\centering
\begin{tabular}{lcc}
\toprule
                                               & QE Task                            & QE-M Task                             \\ \hline
Domain                                         & News                               & Multiple *                            \\
\# sentence pairs                             & 511                                & 2,037                                 \\
\# references                                 & 2                                  & 2                                     \\
\# MT systems                                 & 1                                  & 14                                    \\
\# QE metrics                                 & 10                                 & 10                                    \\
\bottomrule
\multicolumn{3}{l}{\begin{tabular}[c]{@{}l@{}}* Conversation, Ecommerce, News, Social\end{tabular}}    
\end{tabular}
\caption{Statistics of WMT22 Tasks on English--Geman.}
\label{tab:meta_eval_qe_data}
\end{table}

\paragraph{Models and Tools} We use Spearman’s rank correlation coefficient $\rho$ for the calculation of 
automatic QE evaluation, i.e., the correlation function used in Equations \ref{eq:gold_eval_qe}, \ref{eq:ref_eval_qe}, \ref{eq:rank_qe_eval} and \ref{eq:seg_level_eval}. For creating synthetic references, we use a German paraphraser available on Huggingface\footnote{\url{https://huggingface.co/Lelon/t5-german-paraphraser-large}}. We consider different reference-based metrics to see which one is suitable for automatic QE evaluation, which includes:
(1) lexical-based metrics: BLEU \cite{papineni-etal-2002-bleu}, TER \cite{snover-etal-2006-study} and chrF \cite{popovic-2015-chrF}; (2) embedding-based metrics: BERTScore \cite{zhang2019bertscore} and (3) neural-based metrics: BLEURT \cite{sellam-etal-2020-bleurt}, UniTE-MUP \cite{wan-etal-2022-unite}, COMET 22 \cite{rei-etal-2022-comet}, xCOMET XL \cite{guerreiro2023xcomet} and MetricX-23 XL \cite{juraska-etal-2023-metricx}.

\subsection{Quality Estimation with $k$NN}
\paragraph{Dataset} We use the TED talks English--German bitext data from the evaluation campaign IWSLT 2014 \cite{cettolo-etal-2014-report} for training/fine-tuning MT models. The dataset includes 174,443 training sentences, 2,052 validation sentences and 4,698 testing sentences. For evaluation, we use TED test split, and additionally an out-of-domain test set from the WMT22 General task, i.e., the same data as for the automatic QE evaluation experiments. 

For generating the train datastore, we use the train split of TED, i.e., the training data of the MT models. Additionally, we try to use an external, non-train datastore generated using the Europarl dataset \cite{koehn-2005-europarl}. From Europarl, we selected a subset with similar size as the TED training set to rule out the data size factor when comparing the external datastore to the TED datastore. We generate this external datastore similarly to the train datastore, where we perform inference with reference-forced decoding on Europarl using the TED-trained MT models.

\paragraph{MT models} We consider two MT models: a model trained from scratch on TED, and a pretrained DeltaLM model \cite{deltalm} fine-tuned on TED. The model trained from scratch uses the transformer base architecture from the Fairseq library \cite{ott2019fairseq}, with 6 encoder layers, 6 decoder layers and embedding size of 512. Its vocabulary size is 10,112. The fine-tuned DeltaLM model uses the DeltaLM base architecture with 12 encoder layers, 6 decoder layers and embedding size of 768. Its vocabulary size is 250,001. For the fine-tuned DeltaLM model, we build the datastore using only the fine-tuning data (TED), not the whole pretraining data of DeltaLM.

\paragraph{Automatic QE evaluation} We focus on evaluating $k$NN-QE on the segment level. We use our automatic evaluation method described in Section \ref{sec:meta_eval}. We calculate the Spearman correlation between segment-level scores generated by the QE metrics and gold scores generated by the reference-based MetricX-23 \cite{juraska-etal-2023-metricx}, since we find MetricX-23 to be the most robust in QE ranking. Information about the token-level experiments on $k$NN-QE can be found in Appendix A.

\paragraph{Baselines} We use the probability output from the MT model as an unsupervised QE baseline. We take the average of the probability for each token to get the segment-level score. The higher the probability, the better the quality, as it is an indication that the MT model is confident. Since our $k$NN-QE approach is unsupervised, we choose the supervised WMT22 COMET-Kiwi model \cite{rei-etal-2022-cometkiwi} as an upper-bound for the performance. 

\paragraph{Tools} For training/fine-tuning MT models, we use the Fairseq library \cite{ott2019fairseq}. For generating the datastore and retrieving $k$NN samples, we use the $k$NN-box toolkit \cite{zhu2023kNNbox}, which makes use of Faiss \cite{johnson2019billion} for efficient similarity search. For embedding sentences, we use an external model from Huggingface\footnote{\url{https://huggingface.co/sentence-transformers/all-MiniLM-L6-v2}}. Experiments were conducted on an Nvidia TITAN RTX GPU with 25 GB of memory.

\section{Results and Discussion}

\subsection{Automatic QE evaluation}
\paragraph{Overall performance} The performance of difference reference-based metrics on ranking QE submissions on the two WMT22 shared tasks is shown in Table \ref{tab:general_meta_eval}. MetricX-23 XL performs the best on the QE-M Task with 0.939 Spearman correlation to human-based ranking. BLEU performs the best on ranking QE metrics on the QE Task with 0.721 Spearman correlation to human-based ranking. Given these high correlations, we conclude that using reference-based metrics is sufficient for automatically evaluating QE metrics.

\begin{table}[htbp]
\centering
\begin{tabular}{lcc}
\toprule
                                                              & QE Task                                                       & QE-M Task                                                  \\ \hline
BLEU                                                          & \textbf{0.721}                                                & 0.333                                                         \\
TER                                                           & 0.685                                                         & 0.782                                                         \\
chrF                                                          & 0.564                                                         & 0.745                                                         \\
BERTScore                                                     & 0.636                                                         & 0.576                                                         \\
BLEURT                                                        & 0.442                                                         & 0.927                                                         \\
UniTE-MUP                                                     & 0.321                                                         & 0.867                                                         \\
COMET 22                                                      & 0.273                                                         & 0.903                                                         \\
xCOMET XL *                                                   & 0.358                                                         & -                                                             \\
MetricX-23 XL                                                 & 0.261                                                         & \textbf{0.939}                                                \\
\bottomrule
\multicolumn{3}{l}{\begin{tabular}[c]{@{}l@{}}*: xCOMET models are trained on WMT22 \\data except for the News domain, thus only \\ valid to be tested on the QE Task data.\end{tabular}}
\end{tabular}
\caption{Overall performance of different reference-based metrics on ranking QE on two public shared tasks.}
\label{tab:general_meta_eval}
\end{table}

We take a closer look at the correlations on the QE Task. It is quite surprising that BLEU has the highest correlation in QE ranking, since BLEU has recently been shown to have worse evaluation performance than other neural-based metrics \cite{freitag-etal-2022-results}. However, BLEU's ranking correlation is quite low on the QE-M Task data as expected. A similar pattern can be observed where MetricX-23 XL has unexpectedly low performance on the QE Task, but good performance on the QE-M task. We assume that the unexpected ranking performance of the metrics on the QE Task data is due to the narrow scope of QE Task: it considers the output of a single MT model on a single domain. On the other hand, the QE-M Task data is on multiple MT systems output on multiple domains. Therefore, we suspect that the results on the QE-M Task are potentially more generalizable, and that MetricX-23 XL is the best metric for ranking QE metrics. Our following experiment results show evidence that supports this assumption.

\begin{figure}[htbp]
\centering
  \begin{subfigure}[b]{0.4\textwidth}
    \includegraphics[width=\textwidth]{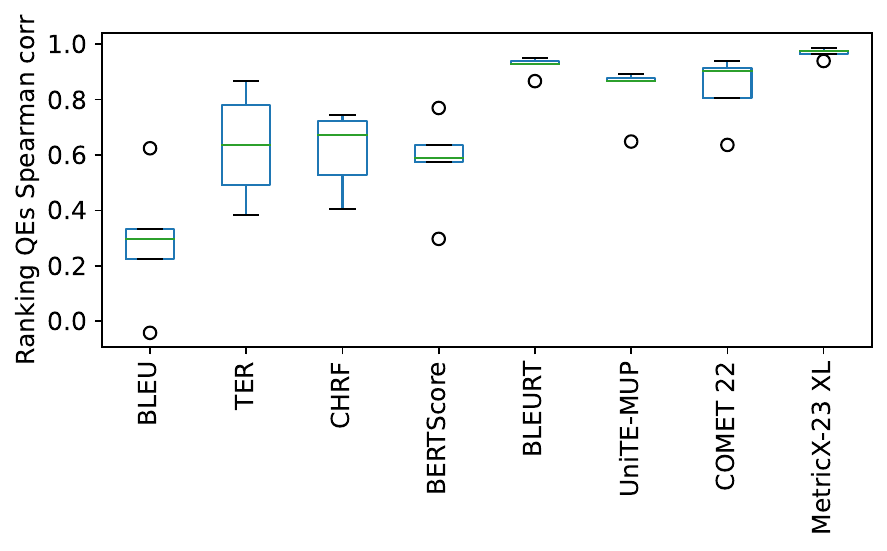}
    \caption{Group by domains in WMT 22 General.}
    \label{fig:corr_gb_domain}
  \end{subfigure}
  \hfill
  \begin{subfigure}[b]{0.4\textwidth}
    \includegraphics[width=\textwidth]{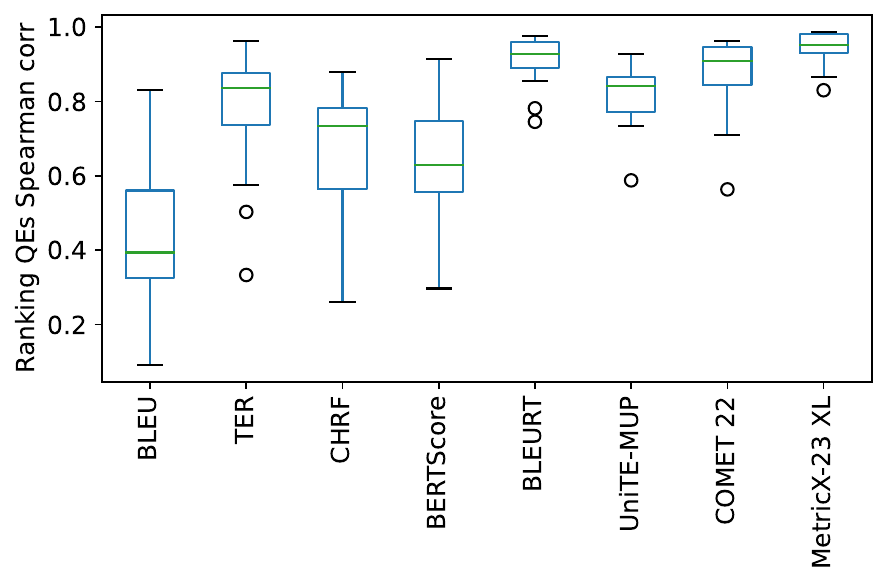}
    \caption{Group by MT systems participated in WMT 22 General.}
    \label{fig:corr_gb_mt}
  \end{subfigure}
  \caption{QE ranking performance across different factors.}
\end{figure}

\paragraph{MetricX-23 XL robustness} We collect the performance of reference-based metrics on QE ranking across different domains and different MT systems' output, as can be seen in Figure \ref{fig:corr_gb_domain} and Figure \ref{fig:corr_gb_mt}, respectively. Generally, the neural-based metrics have better performance than the lexical-based and embedding-based metrics. Their scores are higher and more consistent across different domains and MT systems' output. Among the neural-based metrics, MetricX-23 XL has the best performance in terms of score and consistency.

\begin{figure}[t]
\centering
  \begin{subfigure}[b]{0.49\textwidth}
    \includegraphics[width=\textwidth]{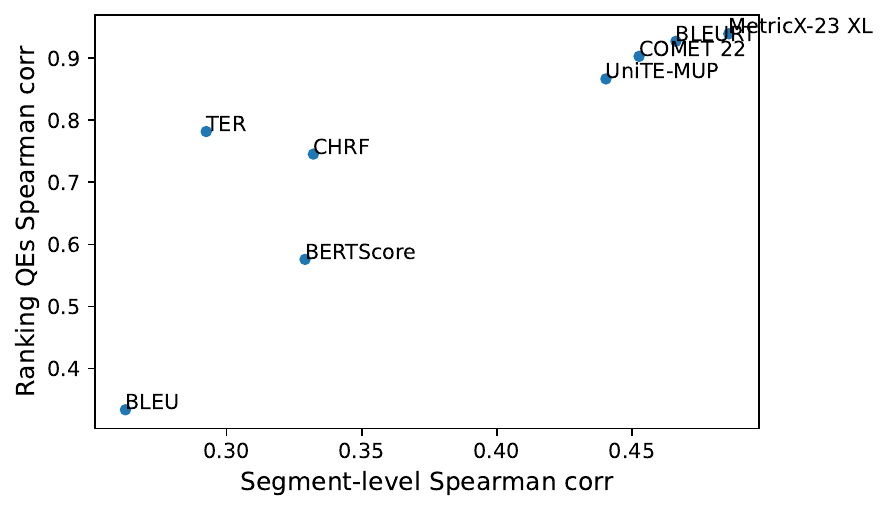}
    \caption{QE-M Task: multiple domains, multiple MT systems.}
    \label{fig:segrank_metricstask}
  \end{subfigure}
  \hfill
  \begin{subfigure}[b]{0.49\textwidth}
    \includegraphics[width=\textwidth]{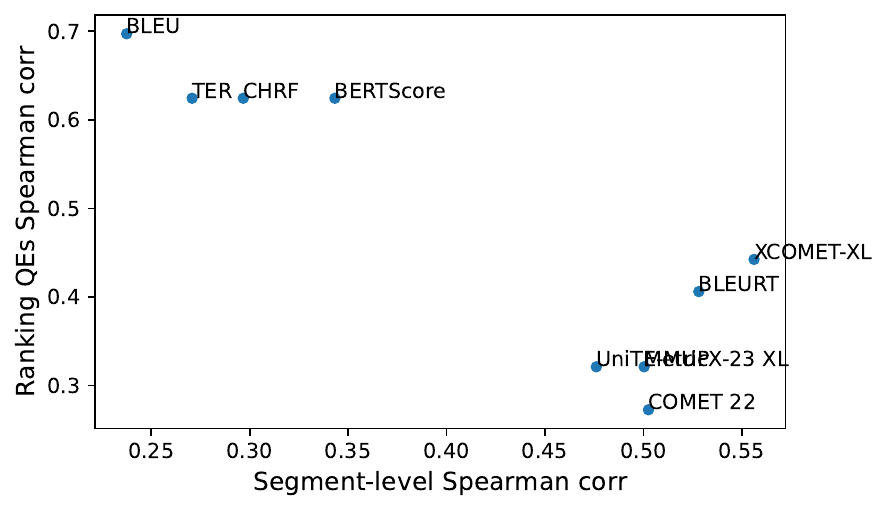}
    \caption{QE Task: single domain, single MT system.}
    \label{fig:segrank_qetask}
  \end{subfigure}
  \caption{Correlation between the performance on evaluating translation segments and the performance on QE ranking.}
  \label{fig:segrank}
\end{figure}

\paragraph{Evaluating segments versus evaluating QE metrics} We investigate whether better performance on evaluating MT segments (Equation \ref{eq:seg_level_eval}) means better performance on evaluating QE metrics (Equation \ref{eq:rank_qe_eval}) for reference-based metrics. Figure \ref{fig:segrank} shows that this is not always the case. For example, in Figure \ref{fig:segrank_metricstask} on the QE-M Task data, TER and chrF have low performance on segment-level evaluation, but have decent performance on ranking QEs. However, both of them are still worse than MetricX-23 XL. In Figure \ref{fig:segrank_qetask} on the QE Task data, the pattern is even more unexpected, where the lexical-based metrics have significantly better performance in QE ranking than the neural-based metrics, while being worse at evaluating segment-level MT output. However, we suspect that this is due to the QE Task data being specific on a single domain and a single MT system's output, thus the result is not representative. The following experiment result supports this assumption. 

\begin{figure}[t]
\centering
  \begin{subfigure}[b]{0.49\textwidth}
    \includegraphics[width=\textwidth]{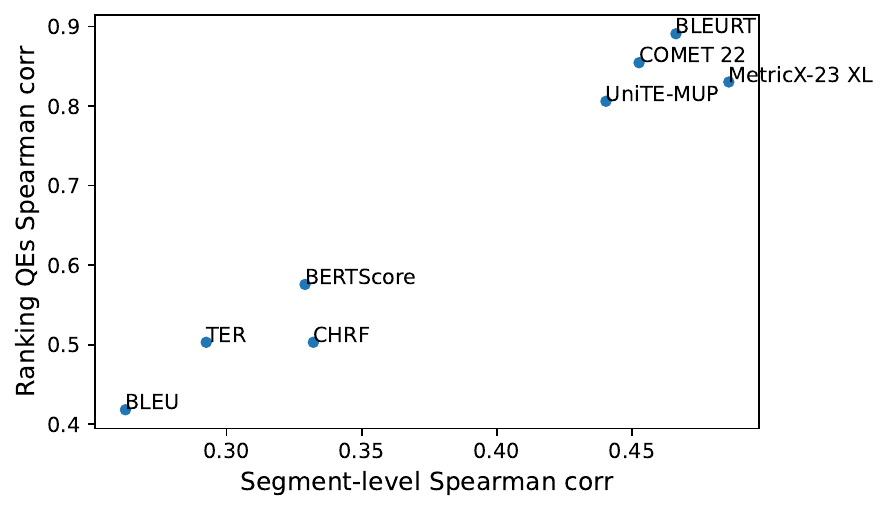}
    \caption{QE-M Task: all domains, single MT system \\(MT system: comet\_bestmbrMT).}
    \label{fig:segrank_all_mbr}
  \end{subfigure}
  \hfill
  \begin{subfigure}[b]{0.49\textwidth}
    \includegraphics[width=\textwidth]{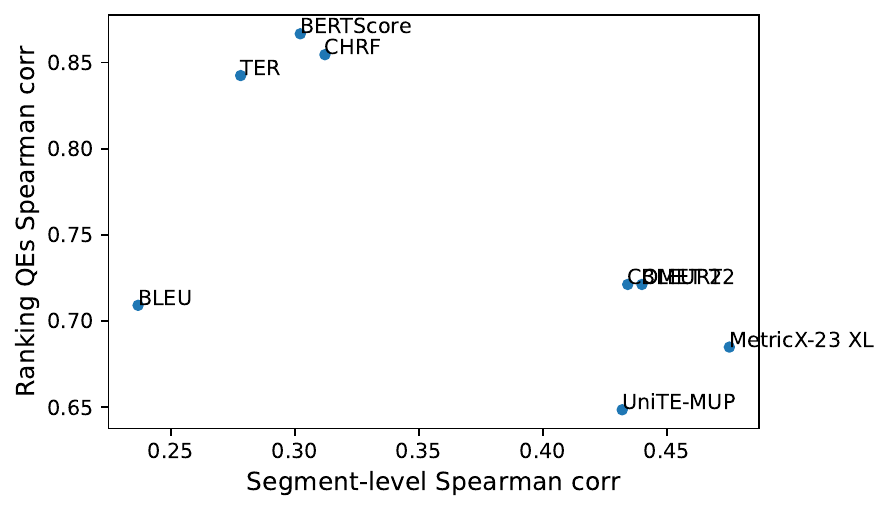}
    \caption{QE-M Task: single domain (Social), single MT system (comet\_bestmbrMT).}
    \label{fig:segrank_social_mbr}
  \end{subfigure}
  \caption{Correlation between the performance on evaluating translation segments and the performance on QE ranking, limited by MT system and domain.}
  \label{fig:segrank_mbr}
\end{figure}

\paragraph{Importance of a broad-ranged test set} We perform the same experiment on segment-level evaluation performance versus QE ranking performance on the QE-M Task, but limit it to a single domain and a single MT system. In Figure \ref{fig:segrank_all_mbr}, on a single MT system output on all domains, the neural-based metrics perform well on both segment-level evaluation and QE ranking as expected. However, on a single MT system output on a single domain (Figure \ref{fig:segrank_social_mbr}), neural-based metrics have worse QE ranking performance than some lexical-based metrics, while still doing well on segment-level evaluation. It can be concluded that reference-based metrics can have unexpected performance when the testing condition is too narrow. Therefore, it is important to perform evaluation on a broad-ranged test set with multiple domains so that we can rely on neural reference-based metrics for QE ranking.

\begin{figure}[htbp]
\centering
\includegraphics[width=0.4\textwidth]{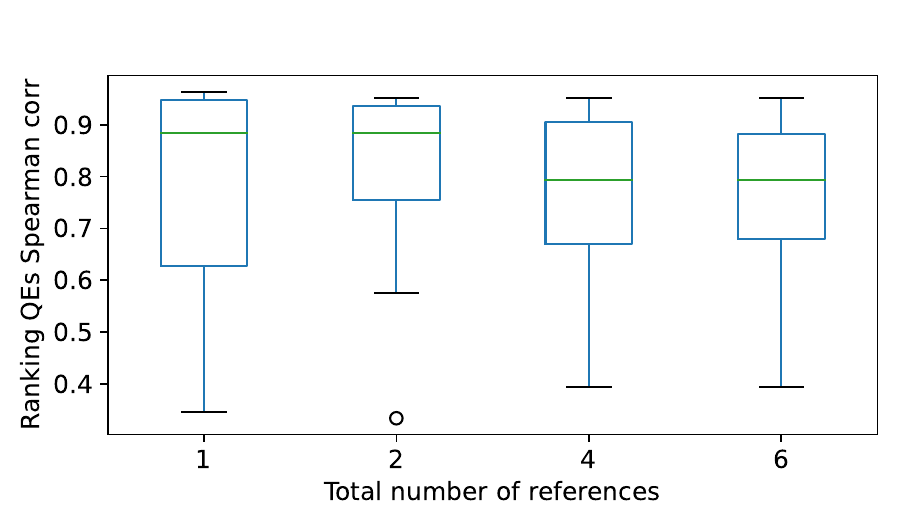}
\caption{QE-M Task: Effect of number of references. The first 2 boxes use human references only, while the last 2 boxes also include synthetic references created by paraphrasing.}
\label{fig:nr_refs}
\end{figure}

\paragraph{Importance of references: quantity and quality} Figure \ref{fig:nr_refs} shows the effect of references on reference-based metrics' performance on QE ranking. Having two human-created references is better than one, showing that increasing the quantity of references helps improve performance. However, adding synthetic references created by paraphrasing decreases the performance to some extent. This shows that it is important to add high-quality references, otherwise it might have the opposite effect of harming the overall performance.

\subsection{Quality Estimation with $k$NN}
We report on the segment-level performance of $k$NN-QE. Experiments on $k$NN-QE performance on the token level can be found in Appendix A.
\begin{table*}[t]
\centering
\begin{tabular}{lllccccc}
\toprule
   &                    &                              & \multicolumn{2}{c}{Transformer Scratch} & \multicolumn{1}{l}{} & \multicolumn{2}{c}{Fine-tuned DeltaLM} \\ \cline{4-5} \cline{7-8} 
   &                    &                              & TED                & WMT22              & \multicolumn{1}{l}{} & TED                & WMT22             \\ \hline
   & \textbf{Baselines} &                              &                    &                    &                      &                    &                   \\
1  & Probability        &                              & 0.535              & 0.525              &                      & 0.462              & 0.423             \\
2  & Supervised QE      &                              & 0.773              & 0.793              &                      & 0.705              & 0.771             \\ \hline
   & \textbf{$k$NN-QE}    &                              &                    &                    &                      &                    &                   \\
3  & TED                & $k$NN token distance $^a$         & 0.650              & 0.623              &                      & 0.575              & 0.438             \\
4  & datastore          & $k$NN sentence similarity $^a$    & 0.570              & 0.553              &                      & 0.527              & 0.398             \\
5  &                    & $k$NN nr. distinct tokens $^b$   & 0.475              & 0.469              &                      & 0.423              & 0.336             \\
6  &                    & $k$NN tokens = output token $^b$ & 0.489              & 0.497              &                      & 0.410              & 0.348             \\
7  &                    & Ensemble $^c$                    & 0.652              & 0.627              &                      & 0.576              & 0.439             \\
   &                    &                              &                    &                    &                      &                    &                   \\
8  & 20\% TED           & $k$NN token distance $^a$         & 0.620              & 0.601              &                      & 0.554              & 0.412             \\
9  & datastore          & $k$NN sentence similarity $^a$    & 0.532              & 0.486              &                      & 0.496              & 0.373             \\
10 &                    & $k$NN nr. distinct tokens $^b$   & 0.498              & 0.494              &                      & 0.407              & 0.373             \\
11 &                    & $k$NN tokens = output token $^b$ & 0.491              & 0.507              &                      & 0.390              & 0.353             \\
12 &                    & Ensemble $^c$                    & 0.622              & 0.604              &                      & 0.555              & 0.413             \\
   &                    &                              &                    &                    &                      &                    &                   \\
13 & Europarl           & $k$NN token distance $^a$         & 0.546              & 0.514              &                      & 0.543              & 0.414             \\
14 & datastore          & $k$NN sentence similarity $^a$    & 0.121              & 0.246              &                      & 0.103              & 0.051             \\
15 & ($\neq$ train)     & $k$NN nr. distinct tokens $^b$   & 0.383              & 0.351              &                      & 0.320              & 0.271             \\
16 &                    & $k$NN tokens = output token $^b$ & 0.437              & 0.465              &                      & 0.384              & 0.335             \\
17 &                    & Ensemble $^c$                    & 0.548              & 0.517              &                      & 0.544              & 0.415             \\

\bottomrule
\multicolumn{8}{l}{\begin{tabular}[c]{@{}l@{}}$^a$: Number of neighbors $k = 1$. \hspace{1cm} $^b$: Number of neighbors $k = 10$.\\$^c$: Ensembling from the other four KNN-QE metrics.\end{tabular}}                              
\end{tabular}
\caption{Overall performance of $k$NN-QE on the segment level.}
\label{tab:$k$NN_qe_overall}
\end{table*}

\paragraph{$k$NN-QE better than MT probability, but worse than supervised QE} The performance of our $k$NN-QE approach is shown in Table \ref{tab:$k$NN_qe_overall}. We consistently observe that the performance of the $k$NN token distance metric is better than the other $k$NN-QE metrics. $k$NN token distance (Row 3) has better performance than the probability baseline (Row 1) by 0.1 increase in Spearman correlation to humans in most cases. However, it still falls behind the supervised QE baseline (Row 2). Ensembling all four $k$NN-QE metrics gives improvement in performance, but not significant, as it is only $\approx$ 0.002 points higher than the performance of the $k$NN token distance metric alone.

\paragraph{Performance diminishes with fine-tuned MT on out-of-domain test set} From Table \ref{tab:$k$NN_qe_overall}, we can see the performance change when moving from the in-domain test set (TED) to out-of-domain test set (WMT22). For the Transformer MT model trained from scratch on TED ("Transformer Scratch"), our approach works for both in-domain and out-of-domain test sets, where it outperforms the probability baseline. On the other hand, for the DeltaLM model fine-tuned on TED, our approach only outperform the probability baseline on the in-domain test set. On the out-of-domain test set, it performs similar or worse than the probability baseline. This is possibly due to the fine-tuned DeltaLM model being pretrained on other data than TED, thus having more knowledge on the out-of-domain test set which is not identifiable if we only use the datastore on TED. It can be concluded that (1) $k$NN-QE works best if we build the datastore using the training data of the model, not only fine-tuning data and (2) with the appropriate training datastore, the approach works for out-of-domain test sets.

\begin{figure}[htbp]
\centering
    \includegraphics[width=0.49\textwidth]{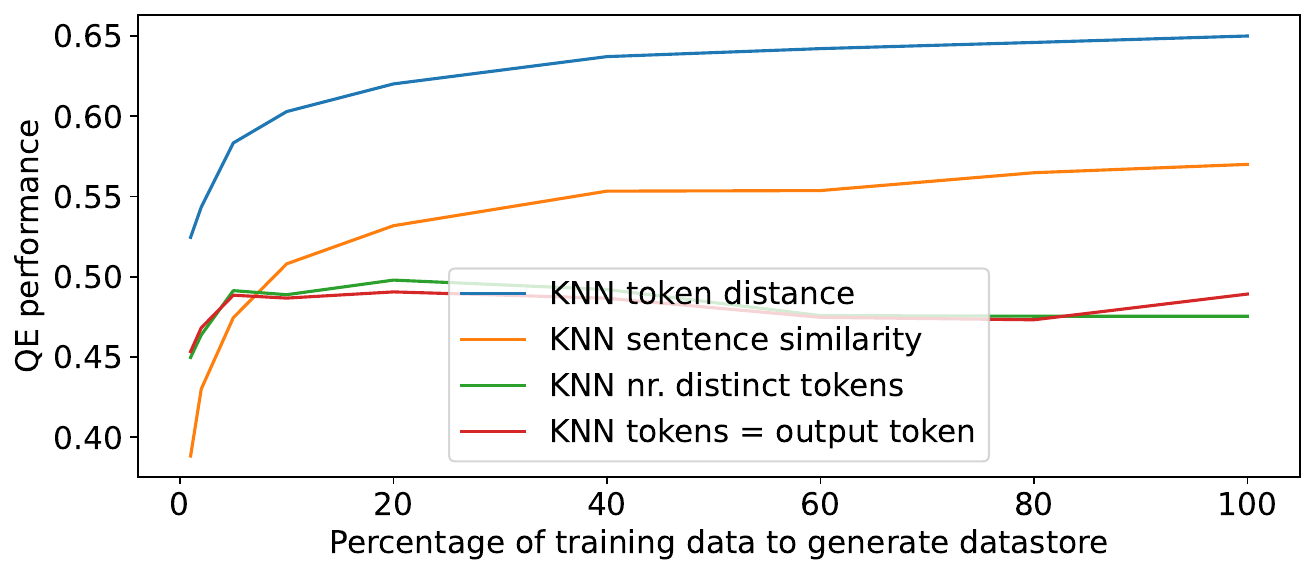}
    \label{}
  \caption{Effect of reduced datastore. Experiment conducted with Transformer Scratch MT model on TED. Similar patterns observed with fine-tuned DeltaLM and WMT 22 data.}
  \label{fig:effect_of_reduction}
\end{figure}

\paragraph{Reducing the datastore has less negative effect than expected} As can be seen in Figure \ref{fig:effect_of_reduction}, generally, the QE performance of the $k$NN token distance and $k$NN sentence similarity metrics increases as the portion of training data used to create the datastore increases. However, the QE performance only increases drastically if we increase the datastore until 20\%, afterward, it starts to flatten. For the other two metrics, the QE performance slightly fluctuates with different datastore sizes. More detailed numbers can be seen in Table \ref{tab:$k$NN_qe_overall}, where the QE performance using 20\% datastore is only worse than using the full TED datastore by $\approx$ 0.03 reduction in Spearman correlation. This is a positive observation, since building a smaller datastore would be more memory-efficient and inference-speed-efficient.

\paragraph{Switching to non-train datastore hurts performance} As can be seen from Table \ref{tab:$k$NN_qe_overall}, changing to the Europarl datastore reduces the performance of $k$NN-QE. This is expected, since a non-train datastore would not be representative of the MT model's knowledge. However, using the Europarl datastore for $k$NN-QE still works to some extent, as the QE correlation to humans is still quite high, at around 0.5 using the $k$NN token distance metrics (Row 13). This is potentially due to the use of forced decoding on reference: in the datastore, we use reference translation as prefix to generate each token, making the generated tokens have higher quality. Thus the closer the inference-time generated tokens are to the high-quality ones in the datastore, the more likely that they also have high quality.

Interestingly, using this same-size but non-train datastore leads to worse performance than using only 20\% of the train datastore, which further strengthens the importance of having a datastore that represents the model's knowledge. 

We also observe that the negative impact of switching to a non-train datastore is less significant for the fine-tuned DeltaLM model than the Transformer model trained from scratch. This is potentially due to DeltaLM's pretraining data containing the same or similar data to the Europarl data, thus the Europarl datastore represents the knowledge of the fine-tuned DeltaLM model to some extent.

\begin{figure}[htbp]
\centering
    \includegraphics[width=0.49\textwidth]{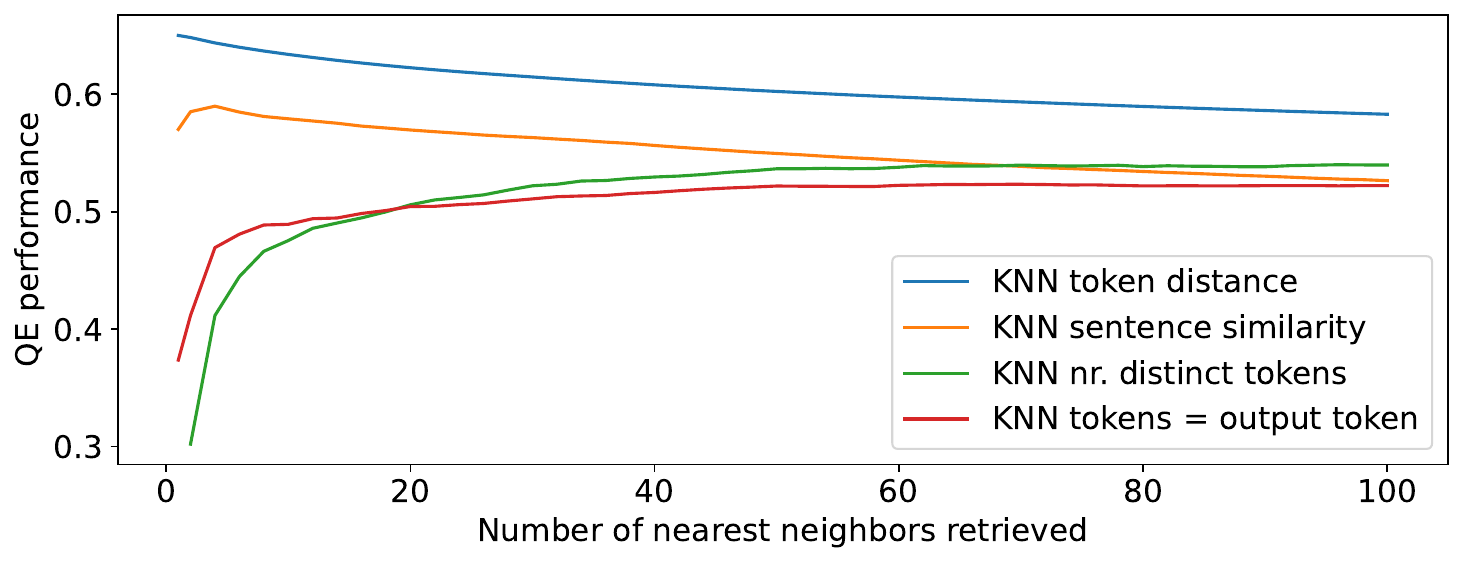}
    \label{}
  \caption{Effect of number of neighbors. Experiment conducted with Transformer Scratch MT on TED. Similar patterns observed with fine-tuned DeltaLM and WMT 22 data.}
  \label{fig:effect_of_k}
\end{figure}

\paragraph{Effect of number of neighbors} As can be seen in Figure \ref{fig:effect_of_k}, $k$NN token distance and $k$NN sentence similarity metrics only need a small number of retrieved neighbors. Their performance decreases as the number of nearest neighbors increases. This is an indication that only the distance of the inference-time generated token to its closest training neighbor matters for these two metrics. However, for the other 2 metrics, i.e., number of distinct $k$NN tokens and number of $k$NN tokens same as model output, the higher the number of neighbors retrieved the better. This is due to these two metrics only comparing the surface-level token output, thus retrieving a small number of neighbors doesn't provide as much information. Based on these observations, we choose the number of neighbors to be $k=1$ for $k$NN token distance and $k$NN sentence similarity metrics and $k=10$ for the other two metrics to report in the main Table \ref{tab:$k$NN_qe_overall}.

Observe that with different numbers of nearest neighbors, the $k$NN token distance metric still performs the best. This means that we can go for this metric in practice with a small number of retrieved neighbors, which benefits the inference speed. Combining the small value of $k=1$ with the reduced 20\% TED datastore, we observe around 19\% increase in inference time when applying $k$NN-QE to the generation process.

\section{Conclusion}
In this paper, we proposed $k$NN-QE -- a model-specific, unsupervised Quality Estimation approach which exploits the information from the MT model's training data. We also propose an automatic QE evaluation method for such model-specific QE approaches, which make use of reference-based metrics. Our experiments show that this 
automatic evaluation method is sufficient, and that the reference-based MetricX-23 XL is the most suitable. Using this automatic QE evaluation method, we found that $k$NN-QE performs better than the MT probability baseline, but still falls behind the supervised QE approach. We also find that our approach works with a small number of retrieved neighbors and a small portion of the training datastore, making it more memory- and time-efficient to be used in practice. For future work, we can explore whether this method is applicable to other types of generative models, such as the currently prominent Large Language Models.


\section*{Acknowledgements}
This work is supported by the Helmholtz Programme-oriented Funding, with project number 46.24.01, project name AI for Language Technologies. It also received partial support from the European Union’s Horizon research and innovation programme under grant agreement No 101135798, project Meetween (My Personal AI Mediator for Virtual MEETtings BetWEEN People). It was partly performed on the HoreKa supercomputer funded by the Ministry of Science, Research and the Arts Baden-Württemberg and by the Federal Ministry of Education and Research.

\bibliography{eamt24}
\bibliographystyle{eamt24}

\begin{table*}[t]
\centering
\begin{tabular}{llcc}
\toprule
Test set & QE method                 & Pearson Correlation & F1-score \\ \hline
TED      & Probability               & 0.07                & 0.08     \\
         & $k$NN token distance $^a$ & 0.21                & 0.27     \\
News     & Probability               & 0.14                & 0.12     \\
         & $k$NN token distance $^a$ & 0.21                & 0.36     \\
\bottomrule
\multicolumn{4}{l}{$^a$: Number of neighbors $k = 1$.}               
\end{tabular}
\caption{$k$NN-QE performance on the token level.}
\label{tab:knn-qe-token}
\end{table*}

\newpage

\section*{Appendix A.\quad $k$NN-QE on token level} \label{app:knn-qe-token}
\subsection*{A.1\quad Experimental Setup}

\paragraph{Motivation for manual evaluation} To evaluate the performance of $k$NN-QE on the token level, we need gold-standard token-level quality labels. On the segment level, we have proposed an automatic QE evaluation method (Section \ref{sec:meta_eval}) by using segment-level quality scores made by reference-based metrics as gold standard instead of human quality scores. In principle, we can do the same for token-level evaluation, by finding a reference-based metric that provides quality labels on the token level. 

However, the performance of reference-based metrics on the token level is usually not as good as on the segment level. For example, the xCOMET metric provides both error-span prediction and segment-level quality scores. Their segment-level quality scores correlate well with human MQM scores, at 0.653 Pearson. Meanwhile, the error-span prediction performance is quite poor, at 0.320 F1 score (although they are still very useful when being aggregated to provide segment scores). This is reasonable, since more fine-grain evaluation tends to be more difficult.

Due to the not-yet-perfect token-level performance of reference-based metrics, we choose not to use them as gold standard to evaluate $k$NN-QE. We instead opt for performing manual annotation on the token level of the MT output to evaluate $k$NN-QE. 

\paragraph{Manually annotated data for evaluation} We manually annotated the MT output on the token level. We annotate 2110 tokens from 100 output sentences on TED data (in-domain test set) and 3503 tokens from 100 output sentences on the News test data (out-of-domain test set).

\paragraph{Metrics} Recall that for each generated subword, $k$NN-QE provides a quality score. To report the performance of $k$NN-QE on the token level, we use two metrics: Pearson Correlation and F1-score. For  Pearson Correlation, we treat the human-annotated labels as continuous scores (0 representing a \textit{BAD} token, 1 representing a \textit{OK} token), and calculate its correlation to the $k$NN-QE scores. For the F1-score, we turn the continuous $k$NN-QE scores into binary labels using a threshold. We choose a threshold that maximizes the F1-score.

\paragraph{Baseline} We compare the performance of $k$NN-QE to an unsupervised baseline using probability output from the MT model. 

\paragraph{Experiment scope} Since it is difficult to perform manual evaluation on a large scale, we limit the scope of our experiment on the token level. In this experiment, we only report on the Transformer model trained from scratch on TED and our best $k$NN-QE metric, i.e., $k$NN token distance. Due to this small scale, we only include the token-level experiment here in the Appendix for more information, rather than including it in the main part of the paper.

\subsection*{A.2\quad Results and Discussion}
As can be seen from Table \ref{tab:knn-qe-token}, our $k$NN-QE outperforms the MT probability baseline. This is an indication that $k$NN-QE also works on the token level. Additionally, we observe that the QE performance is generally better on the out-of-domain test set.

\end{document}